\documentclass[letterpaper, 10 pt, conference]{ieeeconf}  

\IEEEoverridecommandlockouts                              

\usepackage{tabularx, caption, longtable}
\usepackage{lipsum}
\usepackage{lscape}
\usepackage{graphicx}
\usepackage{subcaption}
\usepackage{hyperref}
\usepackage{rotating}
\usepackage{tabularray}
\UseTblrLibrary{booktabs}
\usepackage{pdflscape}
\RequirePackage{xcolor}
\definecolor{tablegray}{HTML}{F2F2F2}
\UseTblrLibrary{varwidth}
\usepackage{ragged2e}

\title{\LARGE \bf
Visual Servoing for Robotic On-Orbit Servicing: A Survey
}

\author{Lina María Amaya-Mejía$^{1,2}$, Mohamed Ghita$^{2}$, Jan Dentler$^{2}$,  Miguel Olivares-Mendez$^{1}$, Carol Martinez$^{1}$
\thanks{$^{1}$Space Robotics (SpaceR) Research Group, Interdisciplinary Centre for Security, Reliability and Trust (SnT), University of Luxembourg, Luxembourg
        {\tt\small  \{lina.amaya},
        {\tt\small  miguel.olivaresmendez},
        {\tt\small  carol.martinezluna\}@uni.lu}}
\thanks{$^{2}$Redwire Space Europe, Luxembourg
        {\tt\small  \{mohamed.ghita},
        {\tt\small  jan.dentler\}@redwirespaceeurope.com}}%
}

\begin{document}
\bstctlcite{IEEEexample:BSTcontrol}

\maketitle
 \thispagestyle{empty}
\pagestyle{empty}

\begin{abstract}


On-orbit servicing (OOS) activities will power the next big step for sustainable exploration and commercialization of space. Developing robotic capabilities for autonomous OOS operations is a priority for the space industry. Visual Servoing (VS) enables robots to achieve the precise manoeuvres needed for critical OOS missions by utilizing visual information for motion control. This article presents an overview of existing VS approaches for autonomous OOS operations with space manipulator systems (SMS). We divide the approaches according to their contribution to the typical phases of a robotic OOS mission: a) Recognition, b) Approach, and c) Contact. 
We also present a discussion on the reviewed VS approaches, identifying current trends. Finally, we highlight the challenges and areas for future research on VS techniques for robotic OOS.

\end{abstract}


\section{INTRODUCTION}

\begin{figure}[ht]
     \centering
     \begin{subfigure}{0.23\textwidth}
         \centering
         \includegraphics[width=\textwidth]{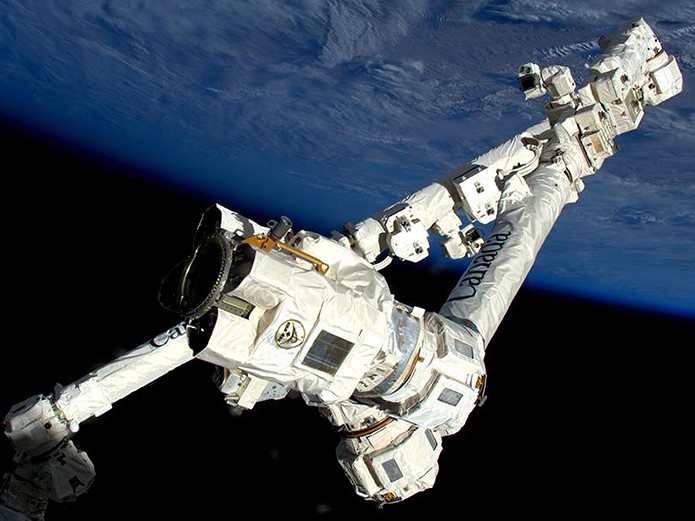}
         \caption{Canadarm2 \cite{canadian-space-agency-2018}}
         \label{fig:canadaarm2}
     \end{subfigure}
     \begin{subfigure}{0.2322\textwidth}
         \centering
         \includegraphics[width=\textwidth]{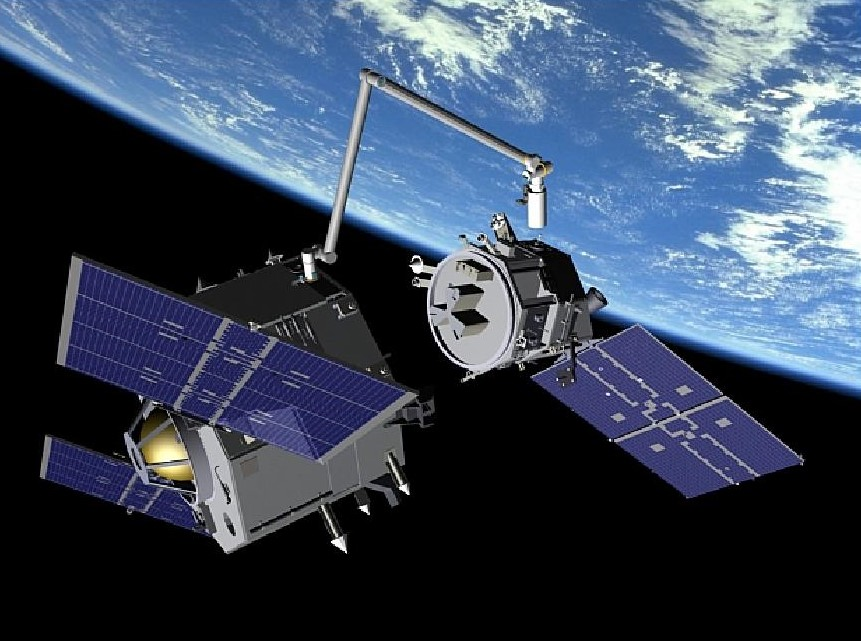}
         \caption{OEDMS \cite{ogilvie2008autonomous}}
         \label{fig:oedms}
     \end{subfigure}
        \caption{SMS on the ISS and satellites for OOS}
\end{figure}

On-orbit servicing (OOS) has the potential to create significant commercial value in the coming years, by extending the spacecraft's lifespan (refuelling, repairing), upgrading it, or redeploying it~\cite{jakhu2017orbit}. 
Space manipulator systems (SMSs) play a crucial role in OOS. An SMS involves a spacecraft equipped with one or more robotic manipulators capable of operating continuously and cost-effectively in environments that may be inaccessible or risky for astronauts, such as geo-synchronous orbit (GEO)~\cite{ma2023advances}.
The use of SMS for OOS began with the Shuttle Remote Manipulator System (SRMS) in 1981. Since then, several space agencies have conducted representative space robotic programs for OOS, both on space shuttles and outside/inside the International Space Station (ISS) and the China Space Station (CSS)~\cite{ma2023advances}. The Canadarm2 and Dextre~\cite{gibbs2002canada} (see Fig. \ref{fig:canadaarm2}) are robotic systems designed for tasks like capturing crew pods, assembling the ISS, and assisting in Extra-Vehicular Activities (EVA). Other ISS-servicing systems include the Japanese Experiment Module Remote Manipulator System (JEMRMS) supporting experiments on the Japanese Experiment Module, and the European Robotic Arm (ERA) with a relocatable base for payload movement and inspection~\cite{flores2014review}. The Chinese Space Station Manipulator system (CSSM) uses two robotic arms for tasks including relocking spacecraft sections, docking assistance, equipment installation, Space Station maintenance, and supporting astronaut EVAs~\cite{wang2019review}.

SMS on satellites has also accomplished OOS in several programs. In 1997, JAXA's ETS-VII~\cite{nishida1999engineering} became the first satellite equipped with a robotic manipulator. In this mission, both automatic and remotely piloted controls were employed to demonstrate rendezvous-docking, along with the teleoperation of the onboard robotic arm. DARPA's Orbital Express Demonstration Manipulator System (OEDMS)~\cite{ogilvie2008autonomous} (see Fig. \ref{fig:oedms}), achieved autonomous capture of a free-flying spacecraft and autonomous transfer of propellant and a battery unit. 
Moreover, OSAM-1~\cite{nasa-2024} was NASA’s On-Orbit Servicing, Assembly \& Manufacturing 1 mission, in which two dexterous arms would execute these tasks using various sensors and vision systems for autonomous real-time relative navigation. However, this mission was cancelled on March 1, 2024 due to technical, cost, and schedule obstacles~\cite{nasa-2024}.

\vspace{-0.5em}

\vspace{-0.5em}

Most of the existing SMS require to be teleoperated by astronauts or ground controllers to function, which is challenging due to communication delays that hinder real-time object manipulation and need highly skilled operators. On the other hand, keeping a human operator in orbit for extended periods or at distant Earth locations is not always logistically, financially, or morally feasible. Hence, OOS missions benefit significantly from autonomous robotic systems capable of making local decisions without human intervention~\cite{moghaddam2021guidance}.
Visuomotor skills can enhance SMS operations, by tracking objects, navigating through complex and dynamic environments, and by improving precision, flexibility and robustness during critical OOS missions. Visuomotor skills are acquired through Visual Servoing (VS) strategies enabling the robot to approach, grasp, and manipulate objects by controlling the robot's relative motion based on visual observations. Nonetheless, several challenges must be addressed to achieve reliable robotic OOS tasks, including the robust identification of targets, precise planning and control for approaching the target, and the mitigation of contact effects.

Therefore, due to the importance and challenges associated with VS techniques in OOS missions, this paper provides a comprehensive review of existing VS methodologies for autonomous OOS involving SMS. The methodologies are categorized based on their primary contributions to the three typical phases of an OOS mission: Recognition, Approach, and Contact. To the best of the author's knowledge, our work represents the first comprehensive overview of existing VS methodologies for robotic OOS missions, identifying current trends and technologies and discussing future directions.
The paper is organized as follows: Section II outlines space environment challenges for VS system design in SMSs. Section III introduces VS principles and approaches. The core of the paper is in Section IV, where a review of existing works on VS approaches for robotic OOS missions is presented according to the target recognition, approach, and contact phases. Section V includes comparison tables of the identified research topics and discusses current challenges and future trends. Finally, Section VI concludes the paper with final remarks.

\section{CHALLENGES}
A key element in enabling autonomous robotic OOS operations is providing the manipulator with the perception that offers awareness of its environment allowing it to interact with it. Cameras stand out as the preferred choice for on-orbit robotic operations due to their advanced technology readiness, reliability, and versatility, particularly when compared to alternative sensors~\cite{ramon2021orbit}. They also provide higher information density and lower costs than GPS or radar systems~\cite{papadopoulos2021robotic}. Cameras can either be fixed on the servicing spacecraft or movable on the manipulator to avoid occlusions during the manipulation task. VS techniques for robotic OOS enable the control of the SMS's motion based on visual feedback~\cite{handbook}, allowing it to respond to changes in real-time. 
Designing the VS system of an SMS performing autonomous operations poses unique challenges due to the unfavourable aspects of the space environment. These include extreme lighting variations, reflectivity of the target's surface, size and shape of the target, ineffectiveness during eclipse, uncertainty, and limited processing power~\cite{sellmaier2010orbit}, which affect the real-time performance of the system. 
Another prominent challenge is micro-gravity, which leads to tumbling object motion. This creates difficulties in testing SMS under representative conditions and in estimating, controlling, and manipulating object movement and position~\cite{papadopoulos2021robotic}. An SMS operating in low Earth orbit (LEO) demands a more complex VS control due to a major number of orbital perturbations. In contrast, missions in geostationary orbit (GEO) allow for a slightly relaxed control with smaller approach velocities given the longer orbital period~\cite{sellmaier2010orbit}. VS techniques for moving targets include solutions that involve reducing time delay through embedded image processing, simplifying algorithms, or focusing on regions of interest~\cite{alabdo2016fpga}. Other approaches include estimating the target state using methods like Kalman filters or image Jacobians. Object tracking approaches have also been proposed to ensure the target always remains within the camera's field of view. 


\section{VISUAL SERVOING}
The aim of VS is control the motion of the robotic system to minimize an error $e(t)$ defined by:

\begin{equation}
\label{eq:vs-error}
    e(t) = s(m(t),a) - s^*
\end{equation}

where $m(t)$ represents image measurements, and, using camera parameters $a$, a vector of $k$ visual features $s(m(t),a)$ is derived. The desired feature set is denoted as $s^*$~\cite{chaumette2006visual}. 
The VS control scheme relates the time variation $\dot{s}$ of the features and the camera spatial velocity $V_c = (v_c, \omega_c)$ with respect to the object. This relationship is given by:

\begin{equation}
\label{eq:vs-feature-vel-relation-partial}
    \dot{s} = L_s V_c + \frac{\partial s}{\partial t}
\end{equation}

where $L_s$ is the interaction matrix constructed based on the selection of the visual features and depth information.  $\frac{\partial s}{\partial t}$ is the time variation of $s$ caused by the object's motion. For a non-moving object, then $\frac{\partial s}{\partial t} = 0$.

To ensure an exponential decoupled decrease of the error $e$ (that is, $\dot{e} = -\lambda e$), the following first-order control law is used:

\begin{equation}
\label{eq:vc-control}
    V_c = -\lambda \hat{L_s^+} e = -\lambda \hat{L_s^+} (s - s^*)
\end{equation}

Where $\lambda$ is a constant gain and $\hat{L_e^+}$ is the approximation of the Moore–Penrose pseudo-inverse of $L_s$.

VS approaches rely on depth information, which is not directly obtained from image measurements for constructing $L_e$. Accurate depth estimation is crucial as it impacts the camera's motion during task execution. It can be derived using geometrical methods, multi-ocular vision, or measured with a depth sensor.

The visual data for a VS scheme can be obtained from a camera on a robot's end-effector (eye-in-hand), fixed in the workspace (eye-to-hand), or both in a hybrid setup. Eye-to-hand offers flexibility in camera placement and panoramic visibility, while eye-in-hand provides more precise control over manipulator motion and the ability to explore the workspace as the robot moves~\cite{hansen2012integrated}.

\subsection{Classical and Advanced Approaches}

Classical VS approaches are classified depending on the definition of the features $s$ guiding the VS control law. In pose-based VS (PBVS), $s$ is a pose defined in the 3D Cartesian space, which must be estimated from image measurements. In image-based VS (IBVS), $s$ is a set of features defined in the 2D image space that are directly obtained from the image~\cite{handbook}.

PBVS has been demonstrated to guarantee better results when used for trajectory planning, but its performance relies on the accuracy of the data, like camera calibration and the model's geometric model. In PBVS, image features can exit the camera's field of view since there is no control over the image plane~\cite{cong2022new}. On the contrary, IBVS has direct control over the feature's trajectory in the image plane, and it does not rely on 3D data, making it more robust towards calibration errors, target modelling, and image noise. However, it can lead to singularities when handling large rotations since there is no control over the 3D trajectory~\cite{palmieri2012comparison}.

Advanced VS methods combine the strengths of classical methods to overcome their limitations, making them suitable for more complex environments. These methods have been designed to decouple some degrees of freedom (DoF) to control them independently using different types of features. The most popular advanced methods are 2.5D VS, which combines both 2D and 3D features to decouple rotational and translational motions, and partitioned IBVS, which only uses features extracted from the image to decouple motions in the Z-axis~\cite{handbook}. 

\subsection{Deep Learning-based Approaches}

VS has been researched and tested on terrestrial applications such as manufacturing, agriculture, and health care~\cite{machkour2022classical}. Versions of the VS previously mentioned use classical computer vision techniques, which are well-established and have been used for many years. They are reliable, transparent and easier to implement. However, they can be affected by noise, occlusions, and lighting conditions. Deep Neural Networks (DNNs) solve these problems by learning patterns from large training sets of labelled images and generalizing these patterns. Modern VS systems employ deep learning approaches since DNNs are invariant to changes in scale, position, illumination, occlusion, and background variations. Still, they are also generally large in size, require high processing power, and extensive training data~\cite{machkour2022classical}. More recent approaches combine depth information from RGB-D cameras with deep learning techniques~\cite{li2022hybrid}\cite{lin2021semantic}.


\section{VISUAL SERVOING FOR ROBOTIC ON-ORBIT SERVICING MISSIONS}
A common robotic OOS mission using VS techniques consists of the following main operational phases: \cite{papadopoulos2021robotic}. 

\begin{enumerate}
    \item Target recognition phase: Initially, the system performs visual identification of the target and feature extraction for estimation of its motion parameters.
    \item Approach phase: Measurements from the previous phase are used as feedback for the control of the robot's end-effector to a position ready to capture the target. 
    \item Contact phase: When the robot's end-effector enters in contact with the target, the SMS must apply the necessary torques and forces such that the target can be controlled. Compliance/impedance controllers are used in complement to VS control.

\end{enumerate}

Executing each of these phases entails specific challenges, including visual identification and parameter estimation of non-cooperative targets, dynamics of SMSs, and handling contact dynamics between the target. The following subsections highlight prominent research advancements on VS for robotic OOS missions, addressing the challenges within each operational phase to enhance flexibility and robustness for forthcoming, more complex robotic OOS missions.

\subsection{Target Recognition Phase} 

A real-time estimation of the motion of the target is essential for planning a collision-free path~\cite{moghaddam2021guidance}. Targets can be divided into cooperative and non-cooperative. A target is cooperative if it is built to be serviced, providing information suitable for pose estimation, such as visual markers, and has a control subsystem that keeps the object in a fixed pose while orbiting. Non-cooperative targets, instead, have uncontrolled attitudes and can be divided into two categories, depending on whether at least geometrical information about their shape and size is available, such as faulty satellites, or they are fully unknown, such as asteroids and most space debris~\cite{opromolla2017review}.

\subsubsection{VS strategies for cooperative targets}

\begin{figure}[ht]
     \centering
     \begin{subfigure}{0.245\textwidth}
         \centering
         \includegraphics[width=\textwidth]{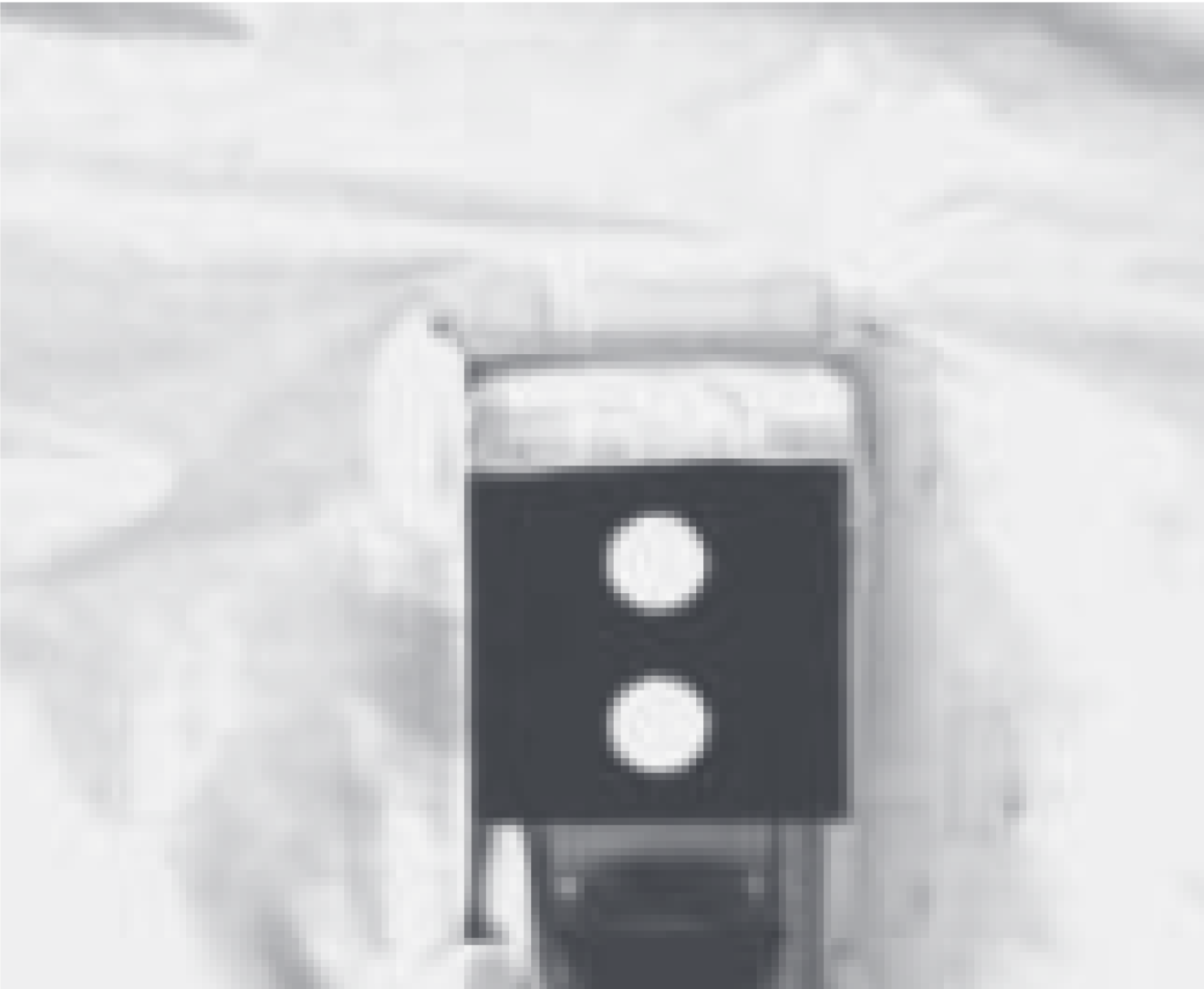}
         \caption{Visual marker on ETS-VII \cite{inaba2000autonomous}}
         \label{fig:ets-markers}
     \end{subfigure}
     \begin{subfigure}{0.203\textwidth}
         \centering
         \includegraphics[width=\textwidth]{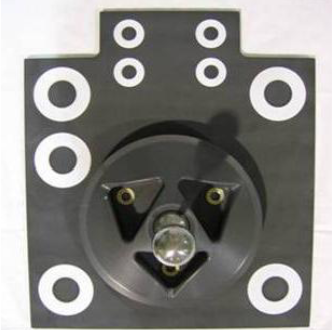}
         \caption{Probe fixture on Orbital Express\cite{ogilvie2008autonomous}}
         \label{fig:oe-markers}
     \end{subfigure}
        \caption{Cooperative visual markers in space}
        \label{fig:cooperative-markers}
\end{figure}

One of the simplest methods for target identification includes the use of fiducial markers~\cite{kalaitzakis2021fiducial} to identify the position of several points on the target~\cite{moghaddam2021guidance}. Spacecrafts are equipped with grapple fixtures and visual markers to be identified and manipulated by an SMS. Cooperative visual perception technologies are currently relatively mature and have been successfully used in many space robots. The ETS-VII~\cite{inaba2000autonomous} robotics experiment utilized VS for the automatic capture of a floating satellite using an SMS. The target's visual detection and pose estimation were facilitated by circular patterns on the handle (see Fig. \ref{fig:ets-markers}). Similarly, the autonomous capture and servicing of a satellite demonstrated by the OEDMS used pre-planned movements integrated with VS~\cite{howard2008advanced} (see Fig. \ref{fig:oe-markers}). DARPA’s Robotic Servicing of Geosynchronous Satellites Program integrates VS methods with a compliance control system to autonomously rendezvous with and perform maintenance and refuelling on large satellites~\cite{wenberg2020development}. In the previously mentioned missions, VS was used for the coarse approach.

A unified open-source framework called OnOrbitROS for space-robotics simulations was introduced in \cite{ramon2023task}. It replicates the primary environmental conditions that space robots may encounter in an OOS scenario. A direct PBVS system was designed to guide the 7-DoF arms of a humanoid robot performing extravehicular operations around the ISS and was evaluated using OnOrbitROS. The system employs eye-to-hand feedback from a camera on the robot's head, and an ArUco marker for target detection and pose estimation. Velocity-based, acceleration-based, and force-based controllers were compared, considering system dynamics and environmental perturbations during the robot's guidance.

\subsubsection{VS strategies for non-cooperative targets}

So far, most SMS has been utilized to service cooperative targets using visual markers and dedicated end-effectors, so the target spacecraft must be equipped with specially designed structures. However, most of actual OOS for faulty satellites are not equipped with fiducial markers~\cite{ma2023advances}. Hence, visual perception for non-cooperative targets is necessary but more challenging due to the unknown characteristics of the target.
 
In \cite{arantes2011rendezvous}, a visual method for real-time pose and motion estimation of a non-cooperative target is presented. It uses a monocular camera system and integrates the target's 3D model and a Kalman Filter (KF) for state estimation. Because of the lack of depth information, visual measurements from a monocular camera can not be completed independently. \cite{xu2006non} presents a comprehensive geometric approach to estimate the target’s geometry and states using stereo vision measurements by the cameras mounted on both the arm and the chaser spacecraft while it rendezvouses with the target. However, stereo matching of the algorithm is time-consuming, resulting in poor real-time performance~\cite{ma2023advances}.

An IBVS system that does not require 3D CAD modelling or fiducial markers was proposed by \cite{shademan2010robust}. By using a modified proportional (P) controller with an uncalibrated Jacobian, this method does not depend on any calibration, neither of robot kinematics nor the vision sensor, making it capable of dealing with unknown targets in unstructured environments. This method was tested on a 7-DoF arm with an eye-hand camera capturing a grapple fixture mockup.

3D vision data has also been introduced for target identification like in \cite{aghili2010fault}, where a computationally efficient noise Adaptive Kalman Filter (AKF) was developed for the motion estimation and prediction of a non-cooperative target satellite in the close-range rendezvous phase. It integrates 3D vision data obtained from a laser camera system (LCS) in harsh lighting conditions and an Iterative Closest Point (ICP) algorithm. The filter receives noisy pose measurements from the LCS onboard the SMS at a close distance and estimates the full states and relative inertia parameters of the target satellite~\cite{papadopoulos2021robotic}. The LCS unit employed was from Neptec Design Group Ltd, which flew successfully onboard the space shuttle Discovery during mission STS-105 to the ISS and subsequently generated real-time 3D imaging data. Later, \cite{aghili2016robust} demonstrated that combining the ICP method with laser scan measurements and inertial measurement unit (IMU) data in a closed loop with an AKF resulted in a robust 6 DoF relative navigation method due to its capability of recovering poses and obtaining dynamic parameters.  

Moreover, visually guided robotic capture of a moving object often requires long-term prediction of the object motion not only for a smooth capture but also because visual feedback may not be continually available~\cite{papadopoulos2021robotic}. 

The IBVS approach outlined in \cite{mithun2018image} exclusively utilizes features in the image space to represent the motion of a tumbling object. This is accomplished by observing that the feature points on the tumbling object follow a circular path around the axis of rotation, and their projection creates an elliptical track in the image plane. This eliminates the need for a direct estimation of the object's motion during the servoing process. The visual features used were the axes and center of the ellipse in the image plane $(a,bx,x_0,y_0)$. In \cite{wang2021research}, a method of predicting the motion state of a moving target in the base coordinate system by hand-eye vision and the position and attitude of the end is proposed. The predicted value is used as the velocity feed-forward, and the position-based visual servo method is used to plan the velocity of the end of the manipulator. To validate their algorithm, they used simulations on an air-bearing platform as well as a real 7-DoF robotic arm with an eye-in-hand configuration to simulate the capture of a grapple fixture.

An approach combining deep learning and 3D image data has been proposed for active space debris removal using PBVS. \cite{lal2021visual} developed a 6-DoF pose detection network trained using YOLOv2 to provide the pose of an object in various lighting conditions and at different angles. A depth camera provided 3D data of the object that was then inputted to the pose detection network. An NVIDIA Jetson Nano was used to test the algorithm due to its good processing capability to handle computer vision algorithms, low power consumption, and small physical footprint, which are desired characteristics for onboard computers. The feasibility of the proposed method was validated on a virtual docking simulation with an industrial robot in an eye-in-hand setup. The algorithm is used to detect a thruster nozzle mockup and detect its pose. A proportional-integral (PI) controller then uses this data to drive the joints of the robotic arm to a desired pose for rendezvous and docking.

\subsection{Approach Phase}
This phase involves the autonomous motion control of the robotic arm based on the feedback data from the previous phase to reach the grasp point on the target. In the case of a moving target, the relative speed between the end-effector and the target must be zero. The challenges in this phase involve avoiding collisions, efficient operation time utilization, mitigating external disturbances, considering SMS dynamics, and preserving continuous line-of-sight to the target's grasping point~\cite{moghaddam2021guidance}.

During this phase, an SMS can operate in two main modes: (a) free-flying mode, in which the on-board AOCS system can keep the orientation of the base spacecraft still, or (b) free-floating mode, in which all spacecraft thrusters are turned off, and the spacecraft translates and rotates in response to manipulator motions~\cite{moghaddam2021guidance}.

\subsubsection{VS in free-flying manipulators}
In \cite{aghili2012prediction}, the approach trajectory for a free-flying manipulator was planned for a tumbling object with uncertain dynamics. This planning involved minimizing a cost function that contained the sum of travel time, distance, object alignment for robotic grasping, and a penalty function related to acceleration. A further advance made by \cite{aghili2021fault} consists of designing an adaptive and fault-tolerant VS system by adding constraints such as visual occlusion and collision avoidance while moving toward the target as fast as possible and ensuring a smooth capture.

In \cite{wenberg2020development}, a hybrid controller was developed to enhance operational effectiveness in the United States Naval Academy's (USNA) Intelligent Space Assembly Robot (ISAR) satellite. This satellite, designed as a remotely-operated orbital assembly testbed, employs robotic arms, 3D camera systems, and sensors for on-orbit assembly and asset maintenance. The hybrid controller combines traditional Jacobian path planning with Visual Servoing (VS) methods to effectively perform these operations. Jacobian path following assumes a well-known environment to plan the end-effector trajectories but struggles in dynamic settings. VS in an eye-in-hand configuration can navigate dynamic obstacles but may follow inefficient trajectories. A hybrid approach can overcome these limitations to enable autonomous task execution under less-than-ideal sensor conditions. The controller was tested on a simulated UR5 6-DoF arm and demonstrated rapid conversion and a small error in the path planning.

\subsubsection{VS in free-floating manipulators}

\begin{figure}[ht]
    \centering\includegraphics[width=0.27\textwidth]{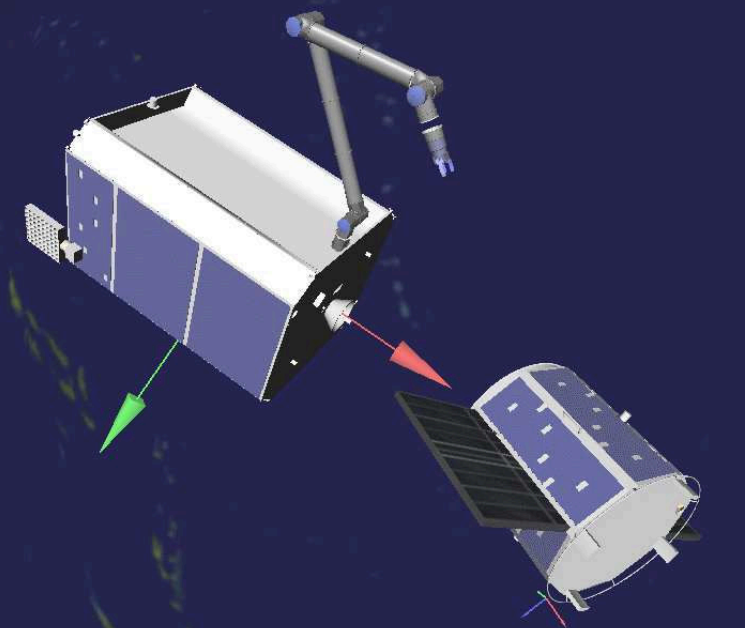}
    \caption{Simulated orbital scenario of an SMS approaching a target satellite \cite{lampariello2013generating}}
    \label{fig:approach}
\end{figure}

In \cite{lampariello2013generating}, a direct single shooting
method was used for the motion planning of a free-floating SMS approaching spinning target (see Fig. \ref{fig:approach}), by integrating robot joint position and velocities constraints, as well as the SMS dynamics. Due to the long computation times involved in the motion planning, a look-up table approach was also presented to provide feasible optimal solutions for a range of spin rates of the target in a useful time.
A reactionless VS control law for a multi-arm SMS is presented in \cite{hafez2017reactionless}. The controller guides the robot's end-effectors to a specific pose while minimizing disturbance to the base satellite's attitude. Using the task function approach and redundancy formulation, tasks are coordinated. The primary task is VS, and a secondary task regulates the base satellite's attitude to zero through a quadratic optimization problem. The effectiveness of the methodology is demonstrated throughout a set of simulations on various multi-arm systems.
In \cite{ramon2022direct}, a direct IBVS algorithm is proposed for the control of a free-floating two-arm manipulator. The algorithm proposed takes into account the relative dynamics of the bodies involved. It relies on images taken from a camera located at the end-effector of a second manipulator. It also integrates an impedance control for the compensation of eventual contact reactions when the end-effector touches and operates the target body.

\subsection{Contact Phase}

\begin{figure}[ht]
    \centering
    \includegraphics[width=0.44\textwidth]{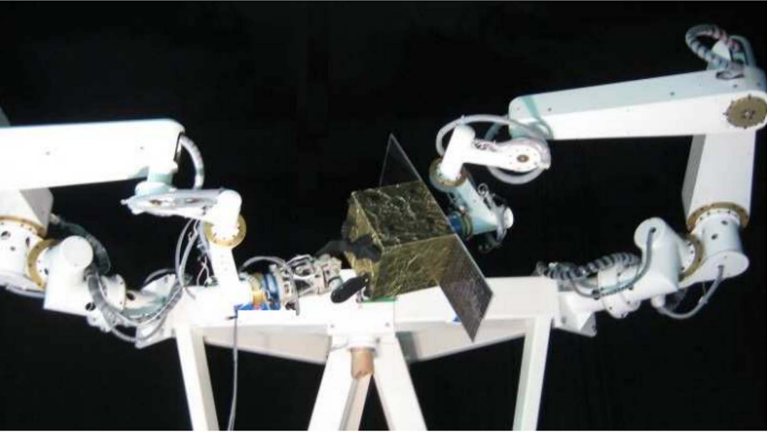}
    \caption{Experimental setup for demonstrating eye-to-hand VS in the contact phase \cite{aghili2023autonomous}}
    \label{fig:contact}
\end{figure}

A fault-tolerant eye-to-hand PBVS to capture and stabilize a tumbling and drifting object is presented in \cite{aghili2023autonomous}, where a switching controller to transition from a pre-grasping to a post-grasping control. The method considers operational and physical constraints, including ensuring a smooth capture, handling target's visual obstructions, and staying within the robot's acceleration, force, and torque limits. For validation, a 6-DoF dual-arm was used, where one of the arms had a satellite mockup and simulated the tumbling motions, while the other arm performs the manipulation task (see Fig. \ref{fig:contact}). 

A significant concern during satellite capture is the potential damage to a manipulator due to twisting or bending. The solution presented in \cite{ma2015hand} involves combining data from two visual systems in eye-in-hand and eye-to-hand configurations to ensure accurate capture and a force/torque sensor on the end-effector to detect contact forces and torques. The base camera extracts image features and guides the manipulator close to the target. Subsequently, the hand-eye camera takes over until the manipulator reaches a pre-insertion position. Finally, impedance control is applied to maintain the compliance of the manipulator.

On the other hand, in \cite{liang2022visual}, PBVS is used in the post-grasping phase for the transposition and docking of an experimental cabin to a core cabin grasped by a robotic arm in free-floating mode. A camera positioned on the core cabin provides eye-to-hand feedback to monitor the relative motion state of the experimental cabin. It is used to control the motion of the robotic arm's joints. The control strategy involves a proportional-–integral–derivative (PID) controller, which dynamically switches between position control mode and velocity control mode based on the distance of the target to the desired parking position before the docking process.

In satellites like the Landsat 7, the fuel ports are covered by a non-rigid multi-layer insulation (MLI) box that must be cut to access the ports. This critical task will be performed by an SMS equipped with a circular blade. The approach in \cite{mahmood2020visual} employs images from an eye-hand camera to monitor and control the cutting process. The method measures the deformation angles on the MLI caused by the blade shaft's pressure as it pushes on the top of the surface. Due to MLI's reflective nature, standard computer vision algorithms struggle in detecting the cutting point. Therefore, A CCTag \cite{kalaitzakis2021fiducial} marker with contrasting colors is attached to the blade for identification. The Canny edge detector extracts marker edges to compute deformation angles, which are then used to estimate the force and engagement depth of the cutting blade. 
An IBVS Proportional (P) controller adjusts the depth of the blade to achieve the desired angles and, consequently, the desired force to achieve a uniform cut along the side. The method was validated through demonstrations using a UR10 6-DoF robotic arm with a force/torque sensor and a spinning blade on its end-effector.

A PBVS in a hybrid eye-camera setup was proposed in \cite{palmieri2023inflatable} for the control of a 2-DoF inflatable manipulator. This architecture allows volume and weight reduction, still maintaining the same payload. The multi-body model takes into account robot dynamics and contact forces with the target. This approach demonstrated through a simulation of an orbital environment that a debris capture operation is possible despite the soft nature of the robot. 


\section{DISCUSSION}
The details of the reviewed VS techniques for robotic OOS missions are summarized in Table \ref{tab:recognition} and \ref{tab:approach}. The former covers aspects of the target recognition phase, whereas the latter presents details related to the approach and contact phases. 

From the review, current trends of VS for robotic OOS missions were identified:

\paragraph{Use of classical VS approaches} 
Most VS approaches for space still make use of classical VS servoing methods, PBVS and IBVS. In particular, a tendency to use PBVS was noted. This VS method is easier to implement when a geometric model of the target is available, or when depth can be obtained from depth/laser sensors or stereo vision to estimate the target's pose. 

\paragraph{Eye-in-hand setup}
An eye-in-hand configuration was mostly preferred among the proposed VS methods since it provides greater accuracy in motion control of the manipulator and does not suffer from occlusions. However, this setup limits the range of robot configurations and provides a limited view of the scene.

\paragraph{Capture of non-cooperative targets}
Researchers have focused their attention mainly on capturing non-cooperative targets. OOS of satellites without visual markers and active debris removal are fundamental for the safety and sustainability of space activities. These targets are tremendously challenging to capture due to the their unknown inertial and dynamic parameters. Hence, most works assume static targets to focus on the development of robust identification methods before dealing with the targets motion states. 

\paragraph{Free-flying mode}
Designing the motion planning and control of a free-floating SMS is a complex procedure, specially when trying to capture moving objects. It is necessary both dynamics of the SMS and the target, as well as the contact dynamics when grasping the target. Additionally, on-ground testing of free-floating is more difficult due to the need of specialized facilities. Therefore, in order to test other methods and technologies, free-flying models are often used.

\paragraph{Use of VS linear controllers}
One of the main advantages of VS techniques is that they are capable of directly mapping visual errors to robot commands using linear controllers, which offer stability, simplicity, and real-time responsiveness. Having a simple linear VS control to approach the SMS's end-effector to the target facilitates researchers to explore switching or hybrid controllers, combing visual feedback with data from force/torque sensors to grasp an object. SMS technologies need advancement to cope with the rising complexity of OOS missions.

A review of the literature reveals the challenges associated with these methods, opening up avenues for development in the following areas:

\setcounter{paragraph}{0}
\paragraph{Use of advanced VS schemes and hybrid eye-hand configurations} 
Classical VS schemes depend strongly on a priori knowledge of the intrinsic and extrinsic camera parameters and their performance can be seriously affected by depth uncertainties and feature loss due to the target's motion. Advanced VS approaches do not require camera calibration or a 3D target model, and are better at keeping the target in the center of the camera's field-of-view. Moreover, a hybrid eye-hand configuration can guarantee precise control of the robot’s end-effector, while keeping a panoramic view of the workspace, combining the advantages of eye-to-hand and eye-in-hand setups.

\paragraph{Development of efficient perception approaches}
Combining RGB and depth sensor modalities enhances perception, as seen in some of the methods reviewed. Depth information provides information in Cartesian space needed for estimating the target's position. Machine learning can further improve object detection, pose estimation, and tracking, overcoming challenges of classical computer vision and extending capabilities to capture non-cooperative targets.

\paragraph{Advanced control strategies for free-floating SMS }
Free-floating SMS have time-varying nonlinear coupling dynamics. The modeling of these dynamics is challenging due to strong coupling between the platform and the manipulator, and unknown disturbances which affect their performance. Hence, there is a need for further exploration of these dynamic characteristics and the development of advanced nonlinear control strategies to achieve precise operations.

\paragraph{Adaptive fault-tolerant controllers }
Linear VS controllers may struggle with highly unstructured environments, obstruction, or large deviations. Adaptive controllers dynamically adjust parameters in real-time, autonomously selecting optimal actions based on the task phase. They enhance dynamic performance, facilitating precise trajectory tracking and smooth execution even in challenging environments or during vision system failures.

\paragraph{Validation in space scenarios }
A major challenge in implementing systems for space applications is transferring the performance of such systems from simulation or ground-based testing to the actual space environment with a known level of reliability. Hardware-in-the-loop is a robust method for  ground emulation of the dynamic behavior of the space environment, including the approach, capture, and docking phases.





\begin{table*}[ht]
    \centering
    \begin{tblr}{
            colspec = {X[0.65] X[0.7] X[0.8] X[0.6] X[0.7] X X[0.8] X[0.7] },
            row{even} = {bg=tablegray},
            cell{1-Z}{1-Z} = {font=\scriptsize\RaggedRight},
        }
    \toprule
    References & VS Scheme & Eye-hand setup & Camera & Visual markers & Features & Target Detection & Target State \\
    \midrule
    \cite{ramon2023task} & PBVS & Eye-to-hand & Virtual & ArUco & 6-DoF pose & ArUco detector & Static (ISS) \\ %
    \cite{shademan2010robust} & IBVS & Eye-in-hand & Mono & Customized pattern & Image coordinates $(x,y)$ & - & Static \\ %
    \cite{aghili2010fault, aghili2012prediction, aghili2016robust, aghili2021fault} &  PBVS & Eye-to-hand & LCS & No & 6 DoF pose & 3D CAD model matching & Tumbling and drifting \\ %
    \cite{mithun2018image} & IBVS & Eye-in-hand & - & No & Ellipse parameters $(a,b,x_0,y_0)$ & - & Spinning \\ %
    \cite{wang2021research} & PBVS & Eye-in-hand & - & No & 6 DoF pose and linear velocity & - & Drifting \\ %
    \cite{lal2021visual} & PBVS & Eye-in-hand & RGB-D & No & 6-DoF pose & Deep learning  & Static \\ %
    \cite{wenberg2020development} & 2.5D VS & Eye-in-hand & Stereo & No & 6-DoF pose + Image's center $(x_0,y_0)$ & - & Static \\  %
    \cite{hafez2017reactionless} & IBVS & Eye-in-hand & - & No & Image coordinates $(x,y)$ & - & Static \\ %
    \cite{ramon2021orbit, ramon2022direct} & IBVS & Eye-in-hand & - & Pattern of points & Image coordinates $(x,y)$ & - & Static \\ %
    \cite{aghili2023autonomous} & PBVS & Eye-to-hand & LCS & No & 6-DoF pose & Point cloud & Tumbling and drifting \\ %
    \cite{ma2015hand} & IBVS & Hybrid & Stereo & No & Image coordinates $(x,y)$ & - & Tumbling \\ %
    \cite{liang2022visual} & PBVS & Eye-to-hand & Ideal model & No & 6-DoF pose & - & Static \\ %
    \cite{mahmood2020visual} & IBVS & Eye-in-hand & RGB & CCTag & Angles in image plane $(\alpha_1, \alpha_2)$ & Canny edge detector & Static \\ %
    \cite{palmieri2023inflatable} & PBVS & Hybrid & Ideal model & No & 6-DoF pose & - & Static\\
    \bottomrule
\end{tblr}
\caption{Details of VS Techniques during the Target Recognition Phase of an OOS mission}
\label{tab:recognition}
\end{table*}




    \begin{table*}[ht]
    \centering
    \begin{tblr}{
            colspec = {X[0.5] X[0.7] X[0.5] X X[0.65] X[0.55] X[0.8] },
            row{even} = {bg=tablegray},
            cell{1-Z}{1-Z} = {font=\scriptsize\RaggedRight}
        }
        \toprule
        References & Base-floating Mode & VS controller & Commands & Robotic arm & Other sensors & Validation \\
        \midrule
        \cite{ramon2023task} & Free-floating & P & Joint velocities, accelerations or forces & Humanoid with 7-DoF arms & No & Software-in-the-loop \\ %
        \cite{shademan2010robust} & Free-flying & P & Joint velocities & 7-DoF & No & Simulation and laboratory \\ %
        \cite{aghili2010fault, aghili2012prediction, aghili2016robust, aghili2021fault} & Free-flying & Adaptive & Camera positions and velocities & 6-DoF & IMU in \cite{aghili2016robust} & CSA testbed \\ %
         \cite{mithun2018image} & Fee-flying & P & Camera velocities & - & No & Simulation and laboratory \\ %
         \cite{wang2021research} & Free-flying & PI & Camera velocities & 3 DoF & No & Simulation and testbed \\  %
         \cite{lal2021visual} & Free-flying & PI & Joint angles & 6-DoF & No & Software-in-the-loop \\ %
         \cite{wenberg2020development} & Free-flying & Switching & Joint velocities & 6-DoF & No & Simulation \\ %
         \cite{hafez2017reactionless} & Free-floating & P with task function approach & Joints velocities & Anthropomorphic multi-arm & No & Simulation \\ %
          \cite{ramon2021orbit, ramon2022direct} & Free-floating & PD & Camera velocities & Anthropomorphic dual-arm & Force & Simulation \\ %
          \cite{aghili2023autonomous} & Free-flying & Adaptive & Camera velocities & 6-DoF & No & CSA testbed \\ %
          \cite{ma2015hand} & Free-floating & P & Joint torques & 6-DoF & Force/torque & Laboratory \\ %
          \cite{liang2022visual} & Free-floating & Switching PID & Joint velocities and angles & 6-DoF & No & Simulation \\ %
          \cite{mahmood2020visual} & Free-flying & P & End-effector translation in Y-axis & 6-DoF arm & Force sensor & Laboratory \\  %
        \cite{palmieri2023inflatable} & Free-flying & P & Joint velocities & Inflatable 2-DoF & Force & Simulation\\

        \bottomrule
    \end{tblr}
    \caption{Details of VS Techniques during the Approach and Contact Phases of an OOS mission }
    \label{tab:approach}
    \end{table*}
    


\section{CONCLUSIONS}

The rapid growth of in-space industrialization, along with commercial OOS capabilities and the increasing complexity of OOS missions, demands improved onboard intelligence and performance for the next generation of SMS. The latter requires urgent breakthroughs in essential technologies to handle diverse visual and dynamic scenarios without human intervention. The development of on-orbit robotic capabilities boosts re-usability, reliability, and safety and eases the execution of proximity operations. Utilizing VS, an SMS can dynamically perceive and respond to its environment in real time, employing visual feedback to control the manipulator's motion. This survey provides a comprehensive summary of advancements in VS for SMS during the target recognition, approach, and contact phases of a robotic OOS mission. It identifies and discusses research trends and ongoing challenges. Additionally, areas requiring further investigation to enhance the safety and reliability of vision-based autonomous on-orbit operations are proposed.


\section*{Acknowledgments}
This work is supported by the Luxembourg National Research Fund Industrial Fellowship grant (N.18075131) and Redwire Space Europe.

\small

\end{document}